\documentclass{article}

\usepackage[utf8]{inputenc} 
\usepackage[T1]{fontenc}    
\usepackage{hyperref}       
\usepackage{url}            
\usepackage{booktabs}       
\usepackage{amsfonts}       
\usepackage{nicefrac}       
\usepackage{microtype}      
\usepackage{graphicx}
\usepackage{algorithmic}
\usepackage{algorithm}
\usepackage{amsmath}
\usepackage{epstopdf}
\title{Robust Learning with Frequency Domain Regularization}

\author{%
	Weiyu Guo, Yidong Ouyang\\
	Information School\\
	Central University of Finance and Economics\\
	\texttt{weiyu.guo@cufe.edu.cn, yidongouyang@gmail.com} \\
}

\begin{document}
\maketitle

\begin{abstract}
   Convolution neural networks have achieved remarkable performance in many tasks of computing vision. However, CNN tends to bias to low frequency components. They prioritize capturing low frequency patterns which lead them fail when suffering from application scenario transformation. While adversarial example implies the model is very sensitive to high frequency perturbations. In this paper, we introduce a new regularization method by constraining the frequency spectra of the filter of the model. Different from band-limit training, our method considers the valid frequency range probably entangles in different layers rather than continuous and trains the valid frequency range end-to-end by backpropagation. We demonstrate the effectiveness of our regularization by (1) defensing to adversarial perturbations; (2) reducing the generalization gap in different architecture; (3) improving the generalization ability in transfer learning scenario without fine-tune.
\end{abstract}

\section{Introduction}
Convolution neural networks (CNNs)\cite{lecunCNN} have achieved remarkable performance in many tasks of computing vision, \emph{e.g.}, object detection\cite{liu2016ssd,yolov3,fasterRCNN}, semantic segmentation\cite{unet}, image captioning\cite{imgCaptioning}, by capturing and representing multi-level features from a huge volume of data. However, existing experiments\cite{poorlyDeep,dodge2017study} demonstrate that CNNs are often with great fragility\cite{foolDeep}. Only injecting minute perturbation, \emph{e.g.}, random noise, contrast change, or blurring, can lead to significant degradation of model performance, \emph{i.e.}, CNN models usually lacks the ability of generalization transfer. 

A variety of explanation of the vulnerability have been proposed, \emph{e.g.}, the limit of the data-sets scale, the distribution of real data is inconsistent with training data, and computational constraints\cite{Bubeck2018AdversarialEF}, which resulting in a variety of coping strategies, such as data augmentation\cite{Yin2019AFP}, adversarial training\cite{Farnia2018GeneralizableAT,Ciss2017ParsevalNI,Miyato2018SpectralNF} and parameter regularization\cite{Wen2016LearningSS,Farnia2018ASA}. In deed, these strategies are propelling the models of CNNs to encoding invariant features as well as neglecting the variable information in the learning phase. In essence, convolutions are a kind of signal processing operations that amplify certain frequencies of the input and attenuate others. This leads us to ask that whether can prompt CNNs to ''remember'' invariant features by explicitly representing certain frequency components of its convolution layers? And, how to find certain frequency components for different layers? In this paper, we show that the answer to the questions leads to some surprising new perspectives on: model robustness and generalization.

The low frequency components in training set are easier to be learned than high frequency components, because the number of low-frequency signals is large but their variation is little. For a finite training set, there exists a valid frequency range, and information beyond the lower bound usually is the bias of data-set, while information beyond the upper bound often is the noise. This phenomenon probably leads to a CNN with common settings always first quickly capturing low frequency components in their dominant, but easily over-fitting when suffering from application scenario transformation. Therefore, we might be able to promote the generalization and convergence performance of CNN models by putting frequency range constraints on convolution layers on learning phase.

The architecture of CNNs is designed to abstract information layer by layer from low to high\cite{Rahaman2018OnTS}. It is generally assumed that, low layers are in charge of extracting low frequency information, such as dots, lines and texture, while high ones are responsible for high frequency information, such as shapes and sketches. Intuitively, we can drive a CNN model to pinpoint the valid frequency range of training set by imposing the low frequency constrains on previous convolution layers while high frequency constrains on the rears. However, due to existing some other factors, e.g., shortcut connection\cite{He2015DeepRL}, learning methods, and sample distribution, the valid frequency range probably entangled in different layers rather than continuous. 

In this paper, we propose a novel frequency domain regularization on convolution layers,  which improves the generalization and convergence performance of CNN models by automatically untangling the spectrum of convolution layers, and navigating the model to the valid frequency range of training set. In a nutshell, our main contributions can be summarized as follows. 
\begin{itemize}
\item An extreme small but valid spectral range for different layers was pinpointed. 
\item A general training approach with frequency domain regularization on convolution layers, for improving the generalization and convergence performance of CNN models. Compared with data augmentation technique and other implicit regularization techniques, our training technique improves the transferability of model.
\item Comprehensive evaluation to investigate the effectiveness of proposed approach, and demonstrate how it can raise the generalization of CNN models.
\end{itemize}

\section{Related work}
Promoting the generalization of models is very important for deep learning. Generally, there are three branches of techniques to achieve the target, \emph{i.e.}, data augmentation, regularization and spectrum analysis.

\textbf{Data augmentation:} The idea of data augmentation\cite{hendrycks2020augmix,Cubuk2018AutoAugmentLA} is common to reduce overfitting on models, which increases the amount of training data using information only in training data. Simple techniques, such as cropping, rotating, flipping, \emph{etc.}, are prevailing in CNN model training, and usually can improve performance on validation a few. However, such simple techniques can not provide any practical defense against adversarial examples\cite{cubuk2017intriguing}, which leads to an emerging direction of data augmentation, i.e., adversarial training\cite{larsen2015autoencoding,bao2017cvae}. Indeed, adversarial training performs unsupervised generation of new samples using GANs\cite{goodfellow2014generative}, which can provide amount of hard examples for training. However, recent studies\cite{Yin2019AFP} demonstrate that adversarial training usually improve robustness to corruptions that are concentrated in the high frequency domain while reducing robustness to corruptions that are concentrated in the low frequency domain.

\textbf{Regularization:} \cite{zhang2016understanding} comprehensively evaluates the performance of explicit and implicit regularization techniques, i.e., dropout\cite{srivastava2014dropout}, weight decay\cite{krogh1992simple,loshchilov2017fixing}, batch normalization\cite{ioffe2015batch}, early stopping. And gives the comments that although regularizers can provide marginal improvement, they seem not to be the fundamental reason for generalization, but the architecture. All of the regularization techniques mentioned above have little effect on preventing the model from quickly fitting random labeled data. 
Sharpness and norms are other perspective for generalization\cite{Neyshabur2017ExploringGI}. There is a tight connection between spectral norm and Lipschitz Continuity, which can be used to flatten minima and bound the generalization error\cite{Bartlett2017SpectrallynormalizedMB,Miyato2018SpectralNF,Farnia2018GeneralizableAT}. Jacobian penalty\cite{Hoffman2019RobustLW} and orthogonality of weights\cite{Prakash2018RePrIT} can also be used for improving generalization. But none of the regularization techniques focus on the transferability of model on unseen domain, nor can they explicit pinpoint the valid range of feature to help the model shield against background and noise.

\textbf{Spectrum analysis:} Indeed, convolution is a common method to extract specific spectrum in signal processing. Inspired by this, there is substantial recent interest in studying the spectral properties of CNNs, with applications to model compression\cite{Chen2016CompressingCN}, speeding up model inference\cite{Mathieu2013FastTO,Vasilache2014FastCN}, memory reduction\cite{Guan2019SpecNetSD}, theoretical understanding of CNN capacity\cite{Rahaman2018OnTS,Tsuzuku2018OnTS}, and eventually, better training methodologies\cite{Yin2019AFP,Fujieda2018WaveletCN,Rippel2015SpectralRF,Liu2018MultilevelWF}. Especially, the works that leverage spectrum properties of CNNs to design better training methodologies are most relevant to this paper. For example, recent study\cite{Jo2017MeasuringTT,Rahaman2018OnTS} find that a CNN model usually is biased towards lower Fourier frequencies while natural images tend to have the bulk of their Fourier spectrum concentrated on the low to mid-range frequencies. From this discovery, some works try to drop the high frequency components from the inputs to improve the generalization of the model, \emph{e.g.}, spectral dropout\cite{Khan2017RegularizationOD} and Band-limited training\cite{dziedzic2019band}. In practice, high-frequency components perhaps are non-robust but highly predictable\cite{Yin2019AFP}. Therefore, although high frequency components contain noise, we do not simply drop them in our work. More in-depth discussion is needed for valid spectral range.

\section{Method}
\begin{figure}[htbp]
	\centering
	\includegraphics[width=.8\textwidth]{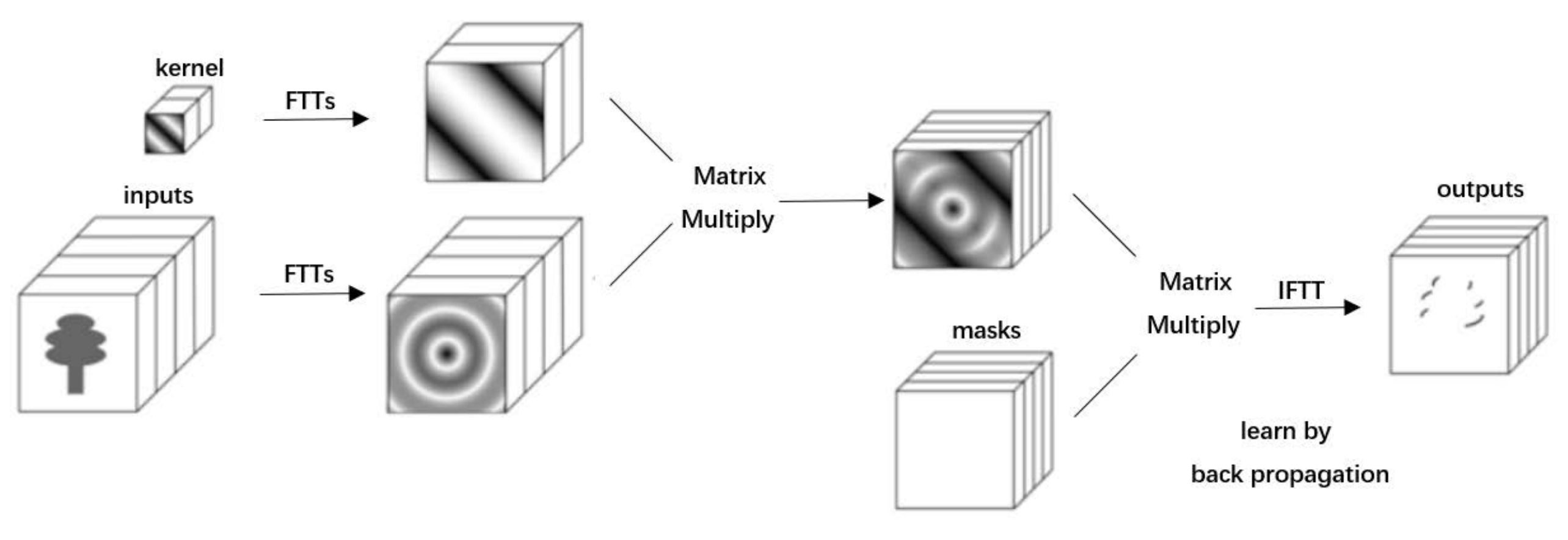}
	\caption{Overview of our method}
	\label{f1}
\end{figure}

In this section, we introduce our regularization method to constrain the frequency spectra of the convolution. The overview of our method is illustrated in Figure \ref{f1}.

\subsection{FFT-based convolution}
\textbf{Fourier transform}

Given a tensor ${t} \in C^{M \times N} $, Fourier transform is used to transform t to the spectral domain.
$$F(x)_{hw}=\frac{1}{\sqrt{MN}}\sum_{m=0}^{M-1}\sum_{n=0}^{N-1}x_{mn}e^{-2\pi i(\frac{mh}{M}+\frac{nw}{N})}
\qquad{\forall}h\ni \lbrace0,1,...,M-1\rbrace,{\forall}w\ni \lbrace0,1,...,N-1\rbrace$$

\textbf{FFT-based convolution}	

The property of frequency analysis ensures that convolution in the spatial domain is equal to element-wise multiplication in the spectral domain. The main intuition of frequency analysis is that an image represented in spatial domain is significant redundancy, while represented in spectral domain can improve filter to feature the specific length-scales and orientations \cite{Rippel2015SpectralRF,Field1987RelationsBT}. Convergence speedup and lower computational cost are additional benefit.

$$x*y=F^{-1}(F_x[w] \cdot F_y[w])$$

$$S[\omega]=F_x[\omega] \cdot F_y[\omega]$$

$S[\omega]$ is called the spectrum of the convolution.

\subsection{Mask design}
Mask design is the key component of our method. Mask helps us pinpoint the valid frequency range entangled between different layers, and using back propagation to update it is the main difference from similar work. 

\textbf{Binarized mask}	

Our regularization try to mask the frequency of background and noise, only maintaining the frequency that is useful for the classification. $M_c[\omega]$ is the mask that limits the spectrum $S[\omega]$.

\textbf{Gradient Computation and Accumulation:}

The gradients of mask are accumulated in real-valued variables, as Algorithm 1.
\begin{algorithm}[!h]
	\caption{Forward and back propagation. $C$ is the cost function for minibatch, $L$ is the number of layers. Quanindicates element-wise multiplication.The function Binarize() specifics how to binarize the masks, and Clip(), how to clip the masks.}
	
	\begin{algorithmic}
		\REQUIRE a minibatch of inputs and targets $(x,y)$.
		\ENSURE updated $Masks M^{t+1}$, Weights $W^{t+1}$.

		\STATE $\lbrace 1.Computing\ the\ masks\ gradients: \rbrace$
		\STATE $\lbrace 1.1.Forward\  propagation: \rbrace$
		\FOR {$k=1$ to $L$}
		\STATE $M_{k}^{b} \leftarrow$ Binarize $\left(M_{k}\right)$
		\STATE $S[\omega]_k^b \leftarrow M_{k}^b \ast S[\omega]_k^b$
		\ENDFOR 
		\STATE $\lbrace 1.2.Backward\  propagation: \rbrace$
		\STATE $\lbrace Please\  note\  that\  the\  gradients\  are\  not\  binary. \rbrace$
		\STATE $automatic\ differentiation\ get\ dS[\omega]$
		\FOR {$k=L$ to $1$}
		\STATE $dM_k \leftarrow dS[\omega]_k \ast S[\omega]_k$
		\STATE $dS[\omega]_k \leftarrow dS[\omega]_k \ast M_k$
		\STATE compute $dx$ and $dy$
		\ENDFOR
		\STATE $\lbrace 2.Accumulating\  the\  parameters\  gradients: \rbrace$
		\FOR {$k=1$ to $L$}
		\STATE $W{k}^{t+1} \leftarrow$ Update $\left(W_{k},\eta,dW_k^t\right)$
		\STATE $M_{k}^{t+1} \leftarrow \operatorname{Clip}\left(\text { Update }\left(M_{k},  \eta, dM_k^t\right),0,1\right)$
		\ENDFOR

	\end{algorithmic}
\end{algorithm}

\section{Experiments}
We demonstrate the effectiveness of our regularization method with various datasets and architecture and compare it with several state-of-art methods. We explore in detail to illustrate the property of our approach.

\subsection{Experimental settings}
\textbf{Datasets} We conduct our experiments with Cifar10\cite{Krizhevsky2009LearningML}, which contains 10 classes, 50000 images for training and 10000 images for testing. 

\textbf{Baseline training} All models use SGD, with momentum set to 0.9. For Cifar dataset, the learning rate is set to 0.01. For Imagenet dataset, the learning rate is set to 0.1. If weight decay and dropout are used, weight decay is set to $10^{-4}$ and the keep-prop is set to 0.9. When training from sketch, Mask is initiated with random numbers from a normal distribution with mean 0.8 and variance 0.2. It means we don’t drop any frequency at beginning, and with the model learning, the accumulated gradient of mask will pinpoint the valid frequency range. While for finetuning, Mask is initiated with random numbers from a normal distribution with mean 0.6 and variance 0.1 to accelerate the learning of mask.
\subsection{Experimental results and analysis}
\begin{figure}[h]
	\centering
	\includegraphics[width=0.8\textwidth]{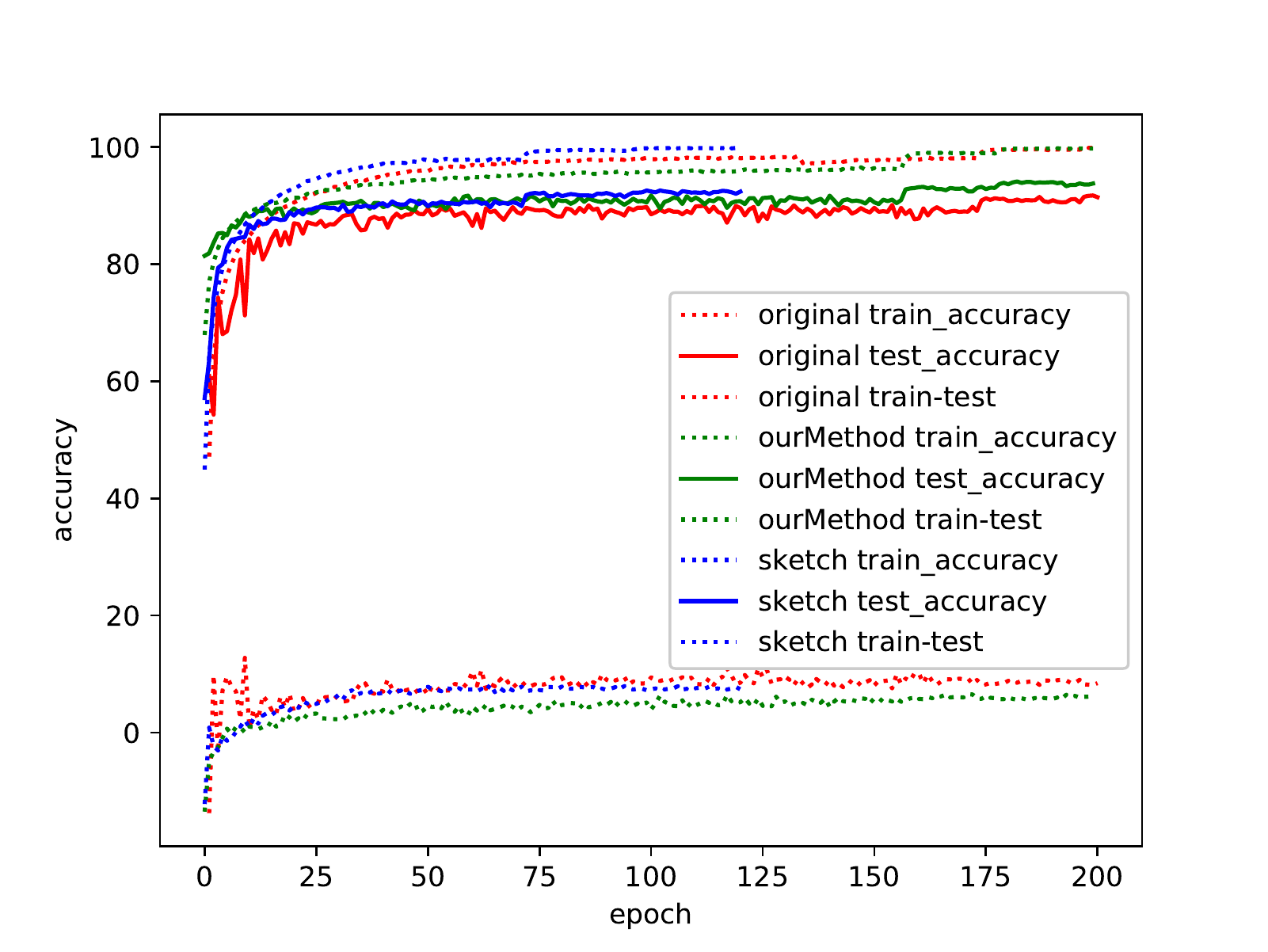}
	\caption{Sketch and finetune}
	\label{result 18}
\end{figure}

\begin{table}[h]
	\begin{center}
	\caption{Summary of test accuracy on Cifar dataset for LeNet architecture. For dropout, DropPath, and SpatialDropout, we trained models with the best  keep\_prob values reported by \cite{Ghiasi2018DropBlockAR}}.
		\begin{tabular}{l|c}
			\toprule
			\hline Lenet baseline & 58.48\\
			\hline Lenet + normalization  & 60.41\\
			\hline Lenet + normalization + random crop + data augmentation & 75.06\\
			\hline Lenet + normalization + random crop + data augmentation + weight decay & 76.23\\
			\hline Lenet + our method + weight decay & 66.7\\
			\hline Lenet + our method + normalization + weight decay & 68.3\\
			\hline Lenet + our method + random crop & 74.0\\
			\hline Lenet + our method + data augmentation & 69.8\\
			\hline \hline Lenet + our method + random drop(0.2)& 62.1\\
			\hline
			\bottomrule
		\end{tabular}
	\label{t1}
	\end{center}
\end{table}

\begin{table}[h]
	\begin{center}
		\caption{Comparison between naturally trained model, Gaussian data augmentation, adversarial training, and our method on clean images and Cifar10-C for resnet-20 architecture.}
		\begin{tabular}{|c|c|c|c|c|}
			\toprule
			\hline  & Clear & Impulse\_noise & Fog & Contrast\\
			\hline Natural &93.5 & 50.436 & 85.14 & 70.858\\
			\hline Gauss & & & & \\
			\hline Adversarial & & & & \\
			\hline Our method & 94.06 & 57.344 & 86.752 & 73.432\\
			\hline
			\bottomrule
		\end{tabular}
	\end{center}
	\label{t2}
\end{table}

\begin{table}[h]
	\begin{center}
		\caption{The percentage of frequency each convolution layer masks for resnet-20 architecture.}
		\begin{tabular}{ll}
			\toprule
			\hline Conv Layer & Mask Percentage \\
			\midrule
			\hline
			layer1conv1.1 & 0.6938 \\
			layer1conv1.2 & 0.5810 \\
			layer1conv2.1 & 0.5213 \\
			layer1conv2.2 & 0.4889 \\
			\hline
			layer2conv1.1 & 0.3354 \\
			layer2conv1.2 & 0.3696 \\
			layer2conv2.1 & 0.3124 \\
			layer2conv2.2 & 0.2644 \\
			\hline
			layer3conv1.1 & 0.2082 \\
			layer3conv1.2 & 0.2621 \\
			layer3conv2.1 & 0.2453 \\
			layer3conv2.2 & 0.1477 \\
			\hline
			layer4conv1.1 & 0.0810 \\
			layer4conv1.2 & 0.0455 \\
			layer4conv2.1 & 0.0337 \\
			layer4conv2.2 & 0.0175 \\
			\hline
			\bottomrule
		\end{tabular}
	\end{center}
	\label{t3}
\end{table}
It was observed that Gaussian data augmentation and adversarial training improve robustness to all noise and many of the blurring corruptions, while degrading robustness to fog and contrast\cite{Yin2019AFP}. Our method has better results against fog, contrast, and impulse noise, which shows that our method alleviates the low frequency brittle caused by adversarial training.
\begin{figure}[htbp]
	\centering
	\includegraphics[width=.8\textwidth]{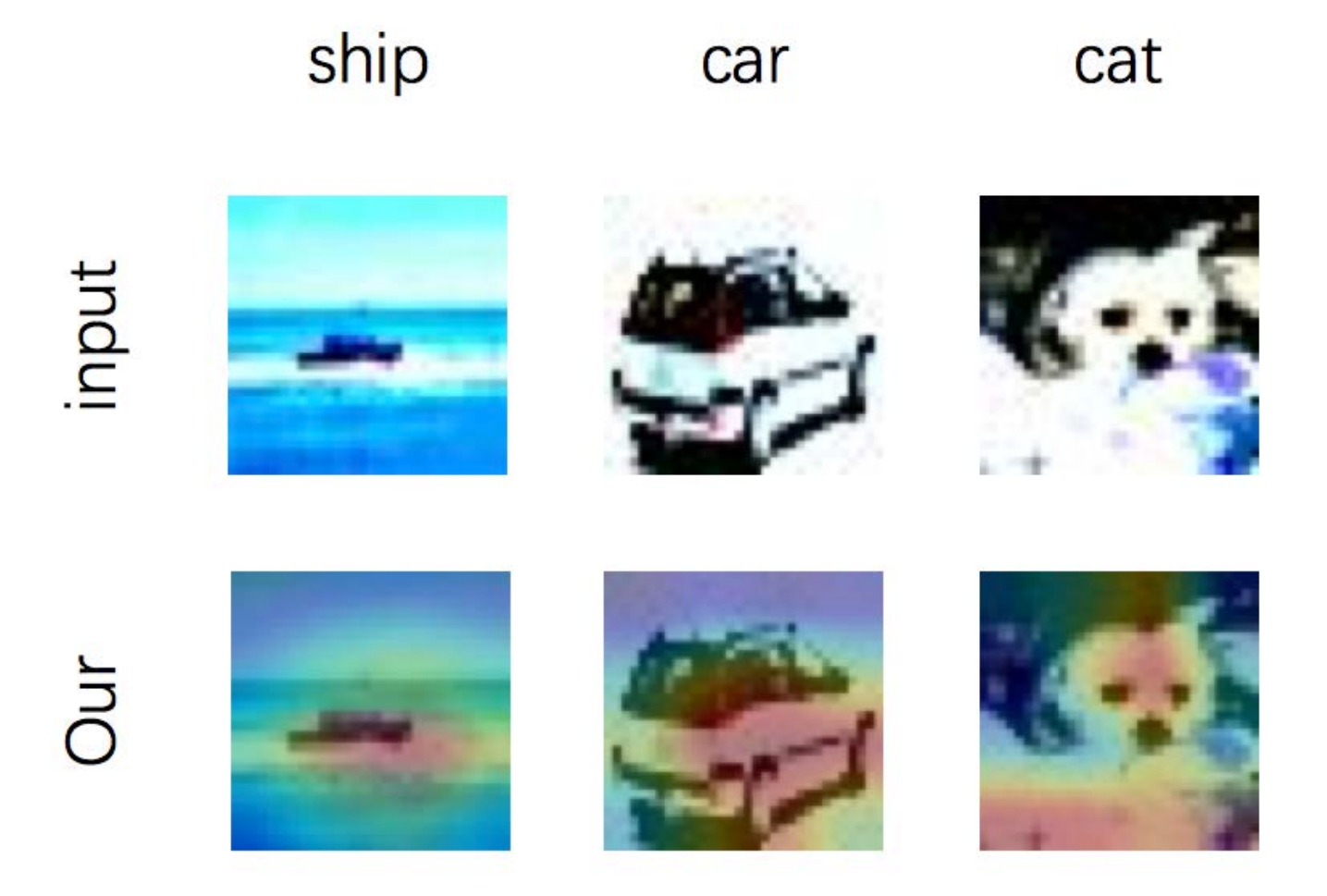}
	\caption{Class activation mapping (CAM)\cite{Zhou2016LearningDF} for resnet-18 model.}
	\label{f2}
\end{figure}

With such amount of activations suppressed in Table \ref{t3} and the CAM illustration in Figure \ref{f2}, our method utilizes the valid frequency range to capture the most important frequency for classification.


\section{Discussion}
\subsection{Binarized mask or continuous mask}
When using Binarized mask in spatial domain, it will have side effect like generating boundary. However, in spectral domain, this kind of side effect is not obvious. It can be seen as a novel way of denoising. Benefitting from the property of spectral domain that represents the feature in an invariant and sparse way, our method can suppress the wrong activation in the spectral domain through binarized mask.

\subsection{Our method with random drop}
Our method pinpoints the valid frequency range for the training set. What if we random drop some frequency after using our method, will the model learn redundancy feature, do not rely on heavily on those frequency and perform better. In the last line of Table 1, we show that it is not a better choice. This verifies that our method pinpoints the valid frequency range on another perspective.

\subsection{Inter-class mask}
If we train the spectral domain mask for each class, it may have better performance and implicit transform among the same category in different datasets. However, in the test time, we need to determine the category of the images before using inter-class mask. So, we may need to change the architecture of the model, which is left to our future work.

\section{Conclusion}
We proposed a novel regularization method in the train time to explicit remove the unimportant frequency. 1) We pinpoint the valid frequency range entangled in different layers. 2) We demonstrate the model trained with our regularization is more robust on unseen data. 

Comparing with Band-limited training\cite{dziedzic2019band} and spectral dropout\cite{Khan2017RegularizationOD}, they do some restriction on spectral domain. Our method differs from them in two aspects: 1) Compared with energy-based compression technique, our method do not drop high frequency component indiscriminately. Our goal is not to minimize the approximation error between masked input and filter with unmasked ones, but to find out the most important frequency for classification and force the model to shield against background and noise. 2) we do not use hyperparameter keep-percentage to determine the threshold for masking. Our method uses back propagation to figure out mask, so our method can be used end-to-end.

Comparing with self-supervised learning strategy \cite{Misra2019SelfSupervisedLO,Pan2020UnsupervisedIA,Xu2020ExploringCR}, our method does not have complex architecture. We try to leverage the transferability of frequency to solve the problem of transferability of model and domain adaptation.

\bibliographystyle{unsrt}
\bibliography{egbib}

\end{document}